\documentclass[letterpaper, 10 pt, journal, twoside]{IEEEtran}
\IEEEoverridecommandlockouts
\usepackage{times}
\usepackage{epsfig}
\usepackage{graphicx}
\usepackage{amsmath}
\usepackage{amssymb}
\usepackage{listings}
\usepackage{alltt}
\usepackage{soul}
\soulregister\cite7
\soulregister\ref7
\soulregister\pageref7
\soulregister{\etal}{7}
\usepackage[linesnumbered,ruled,vlined]{algorithm2e}

\usepackage{cite}
\usepackage[pagebackref=true,breaklinks=true,colorlinks,bookmarks=false]{hyperref}

\newcommand{\etal}{\emph{et al.}}

\newcommand\T{\rule{0pt}{2.1ex}}       % Top strut
\newcommand\B{\rule[-0.9ex]{0pt}{0pt}} % Bottom strut

\begin{document}
    \title{
    Multi-Task Regression-based Learning for Autonomous Unmanned Aerial Vehicle Flight Control within Unstructured Outdoor Environments
}

\author{
    Bruna G. Maciel-Pearson$^{1}$, 
    Samet Ak\c{c}ay$^{1}$, 
    Amir Atapour-Abarghouei$^{1}$, 
    Christopher Holder$^{1}$ and 
    Toby P. Breckon$^{1}$%

\thanks{
    Manuscript received: February, 24, 2019; Revised April, 23, 2019; Accepted July, 1, 2019.
}%Use only for final RAL version

\thanks{
    This paper was recommended for publication by Editor Eric Marchand upon evaluation of the Associate Editor and Reviewers' comments.
    This work was supported by (EPSRC/1750094)
} %Use only for final RAL version

\thanks{
	$^{1}$B.G. Pearson, S.Ak\c{c}ay, A.Atapour, C.J. Holder and T.P.Breckon are with Department of Computer Science, Durham University, Durham, UK (e-mail: \{
	    \href   {mailto:b.g.maciel-pearson@durham.ac.uk}
                {b.g.maciel-pearson},
		\href   {mailto:samet.akcay@durham.ac.uk}
		        {samet.akcay}, 
		\href   {mailto:amir.atapour-abarghouei@durham.ac.uk}
                {amir.atapour-abarghouei}, 
        \href   {mailto:c.j.holder@durham.ac.uk}
                {c.j.holder}
		\href   {mailto:toby.breckon@durham.ac.uk}
		        {toby.breckon}
	\}@durham.ac.uk).
}%

\thanks{Digital Object Identifier (DOI): see top of this page.}
}

\markboth{
    IEEE Robotics and Automation Letters. Preprint Version. Accepted July, 2019
}{
    Maciel-Pearson \MakeLowercase{\etal}: MTRL Autonomous UAV Flight Control within Unstructured Outdoor Environments
} 

\maketitle

\begin{abstract}
    Increased growth in the global Unmanned Aerial Vehicles (UAV) (drone) industry has expanded possibilities for fully autonomous UAV applications. A particular application which has in part motivated this research is the use of UAV in wide area search and surveillance operations in unstructured outdoor environments. The critical issue with such environments is the lack of structured features that could aid in autonomous flight, such as road lines or paths. In this paper, we propose an End-to-End Multi-Task Regression-based Learning approach capable of defining flight commands for navigation and exploration under the forest canopy, regardless of the presence of trails or additional sensors (i.e. GPS). Training and testing are performed using a software in the loop pipeline which allows for a detailed evaluation against state-of-the-art pose estimation techniques. Our extensive experiments demonstrate that our approach excels in performing dense exploration within the required search perimeter, is capable of covering wider search regions, generalises to previously unseen and unexplored environments and outperforms contemporary state-of-the-art techniques. 
\end{abstract}

% \begin{IEEEkeywords}
% IEEE, IEEEtran, journal, \LaTeX, paper, template.
% \end{IEEEkeywords}

\IEEEpeerreviewmaketitle

% \fi
% \IEEEpeerreviewmaketitle

    \section{Introduction}

\begin{figure}
    \centering
    \includegraphics[width=\linewidth]{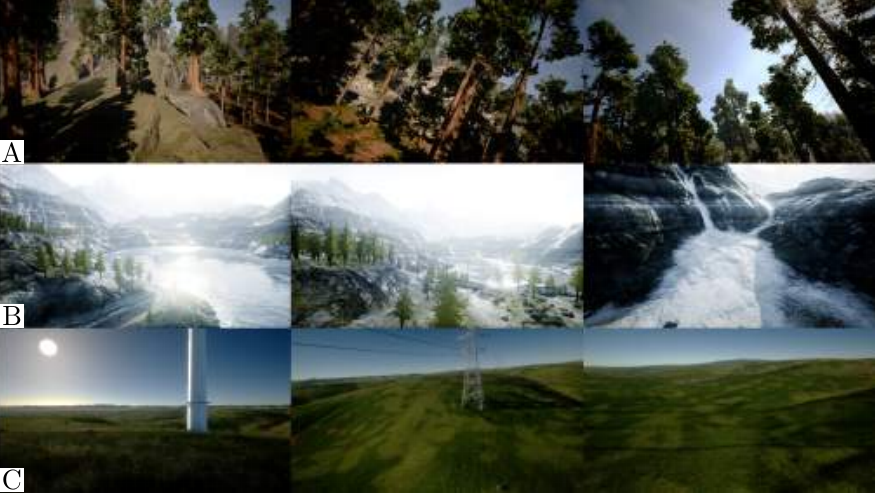}
    \caption{
        Exemplar imagery for autonomous flight and exploration through the AirSim simulator \cite{airsim2017fsr}. Images are from (A) the dense redwood forest, (B) snowy mountain and (C) the wind farm environments.
    }
    \label{fig:testing_set}\vspace{-0.7cm}
    %\todo[inline]{Remove the whitespace around the figure. Fit the figure to column}
\end{figure}

\IEEEPARstart{A}{utonomous} exploration and navigation in unstructured environments remains an unsolved challenge mainly because flights in such areas carry a higher risk of collision and are further aggravated by the power consumption/availability per battery, which is usually limited to less than 20 minutes. In this context, unstructured areas correspond to environments such as disaster areas \cite{adams2011survey}, icebergs \cite{carlson2018adapting}, snowy mountains \cite{karaca2018potential} and forests \cite{Chiara2017Forestry}. These environments tend to have exceedingly variable nature (Fig. \ref{fig:testing_set}) and are often affected by constant changes in local wind conditions. As a result, the mission planning process needs to take into consideration arbitrary, unknown environments and weather conditions \cite{Sebbane2018IntelligentUAV}. Since it is not feasible to preconceive large quantities of unforeseen events that may occur during a mission, an intelligent UAV control system with the capability to generalise to other \textit{domains} is highly desirable. In this context, a new domain can be defined as an area previously unexplored by the UAV, which differs from the original environment on which the deep neural network (DNN) is trained. 

Currently, the existing literature on autonomous flight tends to focus either on mapping the environment, where obstacle avoidance is readily achievable or on deploying models approximated by DNN \cite{Kanellakis2017SurveyTrends}. The former is commonly achieved by techniques such as SLAM or Visual Odometry \cite{perez2018architecture}, which require prior knowledge of the camera intrinsic parameters, while the latter requires a substantial volume of data which is often intractable to obtain \cite{Kanellakis2017SurveyTrends, hentati2018simulation}. 

The focus of our research is navigation within unstructured environments, which is primarily achieved by autonomous exploration under the forest canopy. In this paper, we present a Multi-Task Regression-based Learning (MTRL) approach that allows a UAV to perceive obstacle-free areas while simultaneously predicting the orientation quaternions and positional waypoints in NED (North, East, Down) coordinates, required to explore the environment safely. Due to the nature of our tests, all the experiments are carried out in a virtual environment using the AirSim simulator \cite{shah2018airsim}. As such, our approach uses the Software In The Loop (SITL) \cite{hentati2018simulation} stack, commonly used to simulate the flight control of a single or multi-agent UAV \cite{lamping2018flymaster, hentati2018simulation}. The navigational approach presented in this paper is also independent of the Global Positioning System (GPS), mainly due to the low reliability of GPS signals under dense forest canopy \cite{perez2018architecture}. The proposed learning-based approach utilises a very simple and light-weight network architecture\footnote{Dataset and source code is available at \href{https://github.com/brunapearson/autoUAV-mtrl}{https://github.com/brunapearson/mtrl-auto-uav}} and robustly generalises to new unstructured and unseen environments. 
 
Extensive evaluations point to the superiority of the proposed approach compared to contemporary state-of-the-art techniques \cite{kendall2015posenet, wang2017deepvo, bojarski2016end}. In addition, flight behaviour is also assessed in a SITL environment with ground truth telemetry data gathered during a simulated flight. To the best of our knowledge, this is the first approach to autonomous flight and exploration under the forest canopy without path following or aid of additional sensors.
    \section{Related Work}
Current research on autonomous navigation for UAV can be divided into two groups based on whether path planning or waypoint navigation is the main objective \cite{haralick1994review, Chiara2017Forestry}. 
Path planning requires understanding the environment ahead and it is usually achievable by pre-mapping the environment or specifying a navigational area as the UAV flight takes place \cite{Dey2014VisionFlight, Salami2014UAVAreas, Smolyanskiy2017TowardAwareness,kahn2017plato}, which means the UAV can operate at a constant speed for a set duration in a specific direction \cite{Maciel-Pearson2018ExtendingCanopyb,drews2017aggressive}. Within the existing literature, various end-to-end learning-based approaches have been employed to derive a set of navigational parameters from a given image, allowing for obstacle avoidance \cite{Smolyanskiy2017TowardAwareness,Dey2014VisionFlight,Maciel-Pearson2018ExtendingCanopyb,sadeghi2016cad2rl}. Additionally, the recent advances made in multi-task systems partially focusing on depth estimation \cite{zhou2017unsupervised, ummenhofer2017demon, yin2018geonet, atapour2019veritatem} can also be potentially beneficial towards a successful obstacle avoidance and path planning approach. However, most existing approaches offers only three degrees of freedom navigation, which makes them unsuitable for autonomous flight in unstructured environments, where the UAV may require to change altitude to avoid certain obstacles. As such, varying height estimation and generalisation are fundamental and can best be achieved by defining waypoints.

Waypoints are usually dependant on the the Global Positioning System (GPS) and can be defined before or dynamically generated during the flight. Once the UAV reaches the first waypoint, usually positioned at a short distance away and in an obstacle-free area, the algorithm moves on to defining the next waypoint, and the flight controller makes the necessary adjustments in speed and direction to reach its goal \cite{kaufmann2018deep, mohta2018fast} safely. Alternatively, the flight speed can be estimated by a neural network allowing more generalised and dynamic navigation with six degrees of freedom \cite{jung2018perception,kaufmann2018deep}.
It is important to highlight that navigation using such GPS waypoints is usually limited in environments where the GPS signal is unreliable or unavailable \cite{perez2018architecture}. 
 
Recently, significant results have been achieved by Deep Neural Networks (DNN) in the task of pose estimation based on monocular imagery. In this sense, the use of a Convolutional Neural Network (CNN) \cite{Kanellakis2017SurveyTrends} to learn and to match features, which aids in camera pose estimation, has become popular with the work of Kendall \etal \cite{kendall2015posenet} and more recently the work of Mueller \etal \cite{mueller2018uas}. However, both approaches rely on prior environmental knowledge before yielding an estimation of the camera position. Further enhancements in predicting camera position have been made possible by integrating a Long-Short Term Memory (LSTM) network into the process \cite{wang2017deepvo}. This recurrent neural network uses gates to handle the vanishing gradient problem, which is very common during back-propagation \cite{hochreiter1997long}.

In contrast, our approach uses Multi-Task Regression-based Learning to individually learn the position of waypoints in NED (North, East, Down) coordinates within the scene in addition to learning the rotational quaternions. As a result, the flight controller can dynamically change position and speed based on the output of our resulting multi-task regression network. Such an approach does not rely on GPS readings to navigate, which makes it suitable for operation in weak or denied GPS signal areas. Additionally, it does not require any knowledge of the camera intrinsics and operates with low-resolution images, which makes it ideal for varying payload UAV. 
    \section{Control System Integration}

In this work, the development as well as testing of the proposed approach and the state-of-the-art comparators \cite{bojarski2016end, kendall2015posenet, wang2017deepvo} is performed using the  open-source AirSim simulator \cite{shah2018airsim}. AirSim is built on the Unreal Engine \cite{karis2013real} and offers physically and visually realistic scenarios for data-driven autonomous and intelligent systems. Due to the closeness of the simulated environment and the real world, the control system, which integrates the simulated flight controller into an autonomous navigation approach, will follow the same structure as the control system in a real UAV with on-board processing capabilities. In our work, each approach tested receives two sets of parameters with distinct measurement units. The first is denoted by the NED values noted in meters, whereas the second one is the rotation and orientation of the UAV established in orientation quaternions. Although quaternions are often a standard representation of attitude in graphical engines, particularly for three-dimensional computations \cite{diebel2006representing}, our main motivation for the use of quaternions is attributed to the fact that calculating quaternions requires a significantly smaller amount of memory than calculating rotational matrices, thereby making them more suitable for on-board processing in drones deployed in the real world.

 A quartenion is a hyper complex number of rank 4, commonly used to avoid the inherent geometrical singularity characteristic of the Euler's method \cite{euler1776novi}, which leads to a loss of one degree of freedom in a three-dimensional space. This reduction happens when two of the rotational axes align and lock together \cite{fresk2013full}. Formally, a quaternion $q$ is defined as the sum of a scalar $q_{0}$ and a vector $q=(q_{1},q_{2},q_{3})$:
\begin{equation}
q = q_{0} + q = q_{0} + q_{1}^{\hat{i}} + q_{2}^{\hat{j}} + q_{3}^{\hat{k}},
\label{quaternions}
\end{equation}
 where $q_{0}$, $q_{1}$, $q_{2}$ and $q_{3}$ denote real numbers, and $\hat{i} = (1,0,0)$, $\hat{j} = (0,1,0)$ and $\hat{q} = (0,0,1)$ refer to the fundamental quaternion unit vectors.
  
During simulation, the position $P_{k+1}$ (Eqn. \ref{poseupdade}) of the body at time $k+1$ is updated by integrating the linear velocity (Eqn. \ref{velocity}) and initial position $p_k$, as shown in Eqn. \ref{poseupdade}.
\begin{equation}
    v_{k+1} = v{k}+\frac{a_{k}+a_{k+1}}{2}.dt,
\label{velocity}
\end{equation}
\begin{equation}
    p_{k+1} = p_{k}+v{k} \cdot dt+\frac{1}{2} \cdot a_{k} \cdot dt^{2},
\label{poseupdade}
\end{equation}
where $a$ represents the linear acceleration obtained by applying Newton's second law added to the gravity vector as illustrated in Eqn. \ref{newton2} and $a_k$ is a function of acceleration over time $k$.
\begin{equation}
    a = F_{net}/m+g
\label{newton2}
\end{equation}

The orientation is updated by computing the instantaneous axis $\hat{\omega} = \omega/|\omega|$ through the angle $\alpha_{dt} = |\omega|\cdot dt$, where $\omega$ refers to the angular velocity in the body frame concerning a fixed (world) frame and can be determined by Newton's equations for dynamics.
 
 \begin{figure*}[!t]
    \centering
    \includegraphics[width=\linewidth]{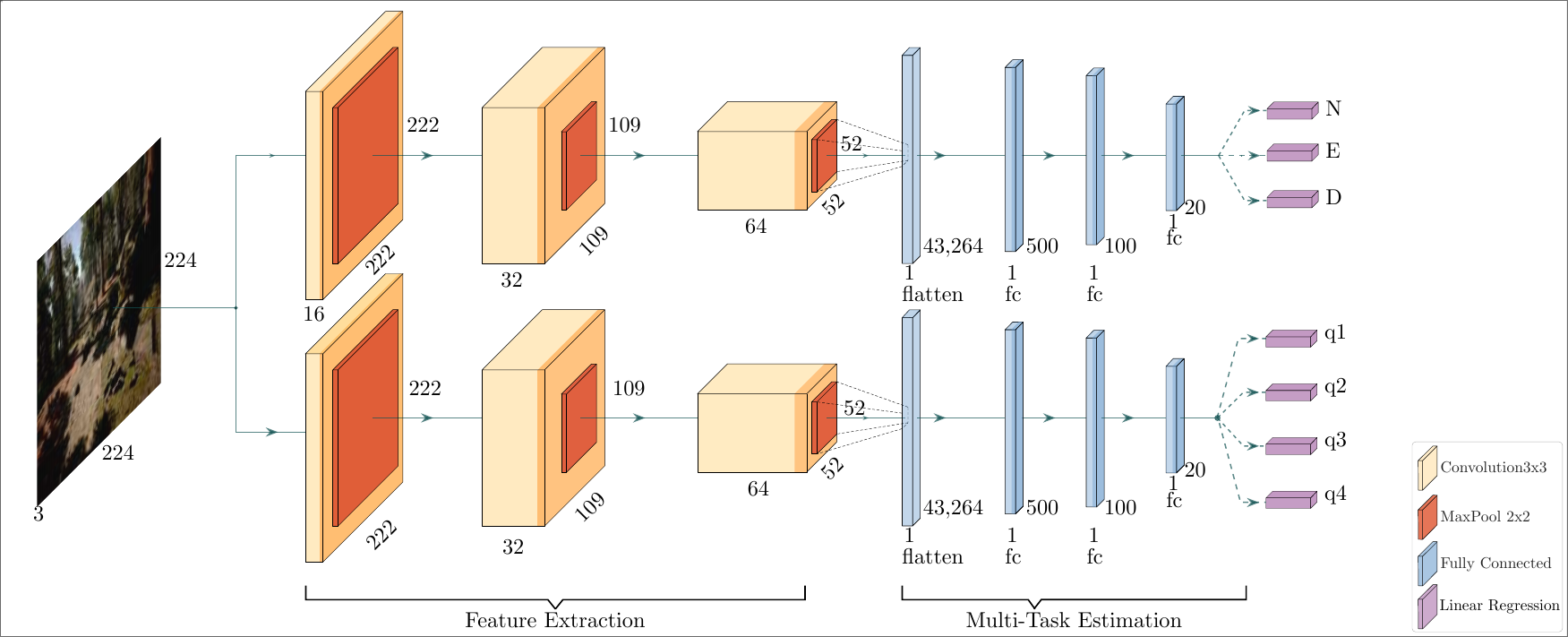}
    \caption{The proposed Multi-Task Regression-based Learning approach. The network predicts 3 positional ($N,E,D$) and 4 rotational values ($q_0$,$q_1$,$q_2$,$q_3$).}
    \label{fig:network}\vspace{-0.4cm}
    %\todo[inline]{Font size not consistent with the main text. Make this PDF. Fit the figure to the linewidth}
\end{figure*}

Flight stability is achieved by combining the rate and attitude control loops at each iteration $\triangle \emph{T}$ (Algorithm \ref{control_loop}). The Rate Control Loop (RCL) has three independent PD (Proportional Derivative) controllers (PD\_roll\_rc,PD\_pitch\_rc,PD\_yaw\_rc) for controlling the body rates ($\dot{\phi}_{D}$, $\dot{\theta}_{D}$ and $\dot{\psi}_D$). The body rates (aka desired body rates) are derived from the target rates ($\phi_{T},\theta_{T},\psi_{T},thrust_{cmd}$), and current rates ($\bar{\phi},\bar{\theta},\bar{\psi}$).  The quartenion values outputted by the network ($model\_output$) are directly fed into the AirSim RCL. This generates the target rates. The Attitude Control Loop (ACL) uses IMU (Inertial Measurement Unit) readings ($\alpha_{x},\alpha_{y},\alpha_{z}$, $\mu_{x},\mu_{y},\mu_{z}$) to estimate the current rate ($\bar{\phi},\bar{\theta},\bar{\psi}$). Thereafter, the motion ($\dot{\phi}_{cmd},\dot{\theta}_{cmd},\dot{\psi}_{cmd}$) commands are generated by the ACL by integrating (PID\_roll\_ac,PID\_pitch\_ac,PID\_yaw\_ac) the desired rates and angular speeds (${\dot{\phi},\dot{\theta},\dot{\psi}}$) acquired from the readings of a 3-axis gyro.

%pseudo code goes here
\begin{algorithm}
\label{control_loop}
\SetAlgoLined
\While{True}{
On each $\triangle \emph{T}$ \;
{$\phi_{T}$,$\theta_{T}$,$\psi_{T}$,$thrust_{cmd}$} $\gets \bold{model\_output}$ \;
{$\dot{\phi}$,$\dot{\theta}$,$\dot{\psi}$} $\gets$ read\_gyro() \;
{$\alpha_{x}$,$\alpha_{y}$,$\alpha_{z}$} $\gets read\_accelerometer()$ \;
{$\mu_{x}$,$\mu_{y}$,$\mu_{z}$} $\gets read\_magnetometer()$ \;
{$\bar{\phi}$,$\bar{\theta}$,$\bar{\psi}$} $\leftarrow attitude$({$\dot{\phi}$,$\dot{\theta}$,$\dot{\psi}$},{$\alpha_{x}$,$\alpha_{y}$,$\alpha_{z}$},{$\mu_{x}$,$\mu_{y}$,$\mu_{z}$}) \;

\tcc{calculate desired body rate}
$\dot{\phi}_{D} \gets PD\_roll\_rc (\phi_{T} - \bar{\phi})$ \;
$\dot{\theta}_{D} \gets PD\_pitch\_rc (\theta_{T} - \bar{\theta})$ \;
$\dot{\psi}_D \gets PD\_yaw\_rc (\psi_{T} - \bar{\psi})$ \;
\tcc{motion commands}
$\dot{\phi}_{cmd} \gets PID\_roll\_ac (\dot{\phi}_{D} - \dot{\phi})$ \;
$\dot{\theta}_{cmd} \gets PID\_pitch\_ac (\dot{\theta}_{D} - \dot{\theta})$ \;
$\dot{\psi}_{cmd} \gets PID\_yaw\_ac (\dot{\psi}_{D} - \dot{\psi})$ \;
\tcc{translate motion commands to PWM signals}
$PWM_{1} \gets thrust_{cmd} - \dot{\psi}_{cmd} + \dot{\phi}_{cmd} + \dot{\theta}_{cmd}$ \;
$PWM_{2} \gets thrust_{cmd} + \dot{\psi}_{cmd} - \dot{\phi}_{cmd} + \dot{\theta}_{cmd}$ \;
$PWM_{3} \gets thrust_{cmd} - \dot{\psi}_{cmd} - \dot{\phi}_{cmd} - \dot{\theta}_{cmd}$ \;
$PWM_{4} \gets thrust_{cmd} + \dot{\psi}_{cmd} + \dot{\phi}_{cmd} - \dot{\theta}_{cmd}$ \;
\tcc{Driving}
$drive\_motor(PWM_{1},PWM_{2},PWM_{3},PWM_{4})$ \;

}%end while loop

\caption{Implementation of Attitude and Rate Control}
\end{algorithm}

\section{Network Architecture}
In our proposed approach, rather than explicitly following a trail, the objective is to identify clear flight areas and predict the flight behaviour while exploring an unknown environment.  As such, to compare the effectiveness of the different approaches described in this work, each technique is required to be adapted in order to produce the same navigational output, which is subsequently mapped into the flight controller.

Originally, the approach by Wang \etal \cite{wang2017deepvo} receives as its input a pair of images and two sets of navigational coordinates. The first is the ($x,y,z$) positional coordinates, and the second is the Euler angles. The distance between the ground truth pose and the predicted pose is minimised during training, resulting in the final output of three positional coordinates ($x,y,z$) and three Euler angles ($pitch$, $roll$ and $heading$). By contrast, during this experiment, we feed to the network the NED positions and orientation quaternion. As such, the output vector is resized to accommodate seven quantities instead of the six initially specified by \cite{wang2017deepvo}. The architecture of the network used in \cite{wang2017deepvo} consists of nine convolutional layers which feed the first LSTM recurrent neural network. This LSTM then supplies its output to a second LSTM network. Each LSTM has 1000 hidden layers. For a detailed description of the architecture, we refer the reader to \cite{wang2017deepvo}.

Similarly, the approach in \cite{bojarski2016end}, which receives one input image and processes three navigational directions, will now output a vector containing seven estimations. Since the approach in \cite{kendall2015posenet} originally receives one input image and outputs the camera's  NED position and orientation in unit quaternion, no changes are required. In all cases, the number of input images originally required for each network is maintained, and all images are resized to $224 \times 224 \times 3$. In addition, all networks receive the coordinates referent to $P_{k+1}$, instead of $P_{k}$. As such, the aim is to predict the next action instead of the current position. Image normalisation is performed according to the specification of each network.\vspace{-0.2cm}

\subsection{Implementation Details}

Our network is based on a Multi-Task Regression-based Learning (MTRL) approach (Figure \ref{fig:network}), where the features learned from the input are shared across two subnetworks to learn the NED position and rotational quaternions. To achieve this, the input image is first resized and pre-processed by applying channel mean subtraction. Next, the normalised image is shared between two identical subnetworks, each with a convolutional layer $(222 \times 222 \times 16)$, followed by a max polling layer $(111 \times 111 \times 16)$, a second convolutional layer $(109 \times 109 \times 32)$, followed by a second max pooling layer $(54 \times 54 \times 32)$, a third convolutional layer $(52 \times 52 \times 64)$, a third max pooling layer $(26 \times 26 \times 64)$, in which the output is flattened $(43264)$ and fed into a dense layer $(500)$, where dropout is applied. For the positional branch, we apply a dropout of $0.5$ while none is applied to the rotational branch. The output is then fed into a second dense layer, to which we apply a dropout of 0.25 only to the rotational branch. The third dense layer (20) performs the linear regression and outputs the estimation for each task.  Our cost function $\theta$ minimises the Euclidean distance between the predicted output $\hat{x}$ and the ground truth $x$. A scalar factor $N$ with assigned value of $0.1$ is inserted to reduce the difference between the positional and rotational errors.\vspace{-0.2cm}

\begin{equation}
\theta = argmin \frac{1}{N} \sum_{i=1} ||{\hat{x} - x}||_2
\end{equation}
    \section{Experimental Setup}

Training is performed using adaptive moment estimation \cite{kingma2014adam} with a learning rate of $0.0001$ and default  parameters following those provided in the original paper ($\beta_1=0.9, \beta_2=0.999$) \cite{kingma2014adam} using a GTX 1080Ti using 64,000 frames for training data and 17,674 frames for validation. During the test, AirSim is set up to use the GPU while all approaches are tested using an Intel Xeon processor.\vspace{-0.1cm}

\subsection{Data Preparation} 
Data is obtained by manually flying the UAV through the Redwood Forest environment using a FrSky Taranis (Plus) Digital Telemetry Radio System. In total, 81,674 frames were captured together with the flight telemetry comprising of flights under and above the forest canopy, navigation inside caves and over river beds, lakes and mountains.\vspace{-0.1cm}

\subsection{Evaluation Criteria}
Our evaluation is presented across five metrics: repetition, generalisation, flight behaviour, distance travelled and reliability.

In search and exploration scenarios, exploring as many routes as possible is often the primary objective, which is why we investigate the capability of different approaches in alternating between different paths rather than continuously repeating the same route. Additionally, we aim to evaluate the capability of various methods to generalise to unseen environments. During flight behaviour analysis, we observe wherever the UAV is flying or hovering and note the flight duration and the distance traversed during the mission. Here, the goal is to identify the method capable of traversing greater distances in a shorter period. Finally, the reliability of the flight mission is measured by considering the distance the UAV can fly without any intervention. 
    \section{Results}
In this section, we evaluate the performance of the proposed approach in forecasting the next position and rotation of the UAV in a given environment.\vspace{-0.1cm}

\subsection{Repetition}
\begin{figure}
    \centering
    \includegraphics[width=\linewidth]{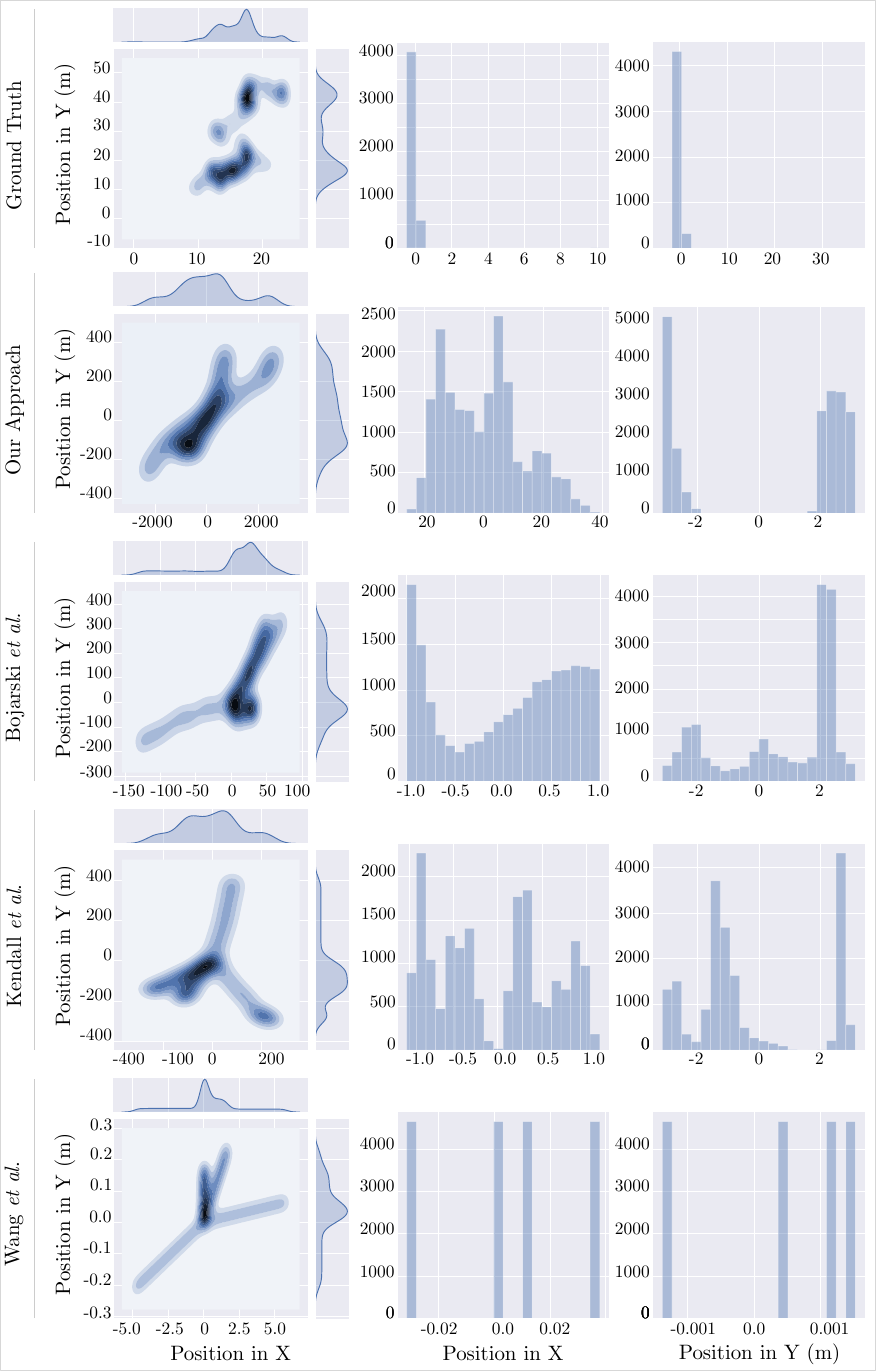}
    \vspace{-0.5cm}
    \caption{2D representation of flights using the test set. Left image shows the combined routes of four flights which are then superposed. Middle and right images show waypoint distributions for the pair ($x,y$) directions, while flying in the default mode of $1m/s$.}
    \label{fig:gt_dist}\vspace{-0.3cm}
    %\todo[inline]{Fonts are too small. Convert the figure to PDF}
\end{figure}

Our first set of tests assess the approach's behaviour when processing the test data (Figure \ref{fig:gt_dist}). For simplicity, we shall define motion in the $N$ coordinate as $y$ and $E$ coordinate as $x$, while $D$ will be depicted as $z$. Here, the flight commands are predicted but not sent to the flight controller. The objective is to observe the navigational patterns of each method and to compare the distribution of the predicted positional values for $x$ and $y$ directions. In order to create the flight pattern shown on the left (Figure \ref{fig:gt_dist}), we add the predicted values of ($x_{k+1},y_{k+1},z_{k+1}$) to the current position ($x_k,y_k,z_k$), just as it would be in the flight controller, using a frame rate of 20fps.

The test set used to compare each network has a high density of repetitive positional values (dark blue areas) distributed over a small area (Figure \ref{fig:gt_dist}). Because this test set mostly contains scenes of low mobility and hovering behaviour, the endeavour is to observe if any of the approaches are capable of understanding the difference between hovering behaviour and slow motion. As illustrated in Figure \ref{fig:gt_dist}, the approaches of Bojarski \etal \cite{bojarski2016end} and Kendal \etal \cite{kendall2015posenet} perform better at slow motion, given that most predicted positional values in $x$ are in the range of $-1.0$ to $1.0$. Similarly, the approach of Wang \etal  \cite{wang2017deepvo} showed a lower variance of predicted values for the $x$ direction (range between $-0.02$ to $0.02$). In contrast, the proposed approach covers more exploratory ground due to significantly higher predicted positional values (range of $-20$ to $40$) as illustrated in Figure \ref{fig:gt_dist}.

 Here, we also evaluate the consistency of each model by superposing each one of the resulting routes after four iterations (Figure \ref{fig:gt_dist}) and although none repeated the same route, the proposed approach shows greater consistency of decision when the same set of images are presented, as evidenced by the overlap/common ground (dark blue) areas in Figure \ref{fig:gt_dist}. The approach of Bojarski \etal \cite{bojarski2016end} predicts similar positional values during three out of four iterations, while the approaches of Kendal \etal \cite{kendall2015posenet} and Wang \etal  \cite{wang2017deepvo} choose different directions at every iteration.

\begin{table}
\small
\caption{Performance of each approach during autonomous flight in the forest, snowy mountain and plain field environments. Reliability is the percentage of the distance travelled without interventions $NI$.}
\label{table:1}
\vspace{-0.5cm}
\begin{center}
\resizebox{\columnwidth}{!}{
\begin{tabular}{|l|c|c|c|c|c|}
\hline
Method  & NI & Reliability & Duration & Distance (m) & Behaviour \T\B \\
\hline\hline
\multicolumn{6}{|c|}{Dense Forest} \T\B \\
\hline

Bojarski \etal \cite{bojarski2016end}& 7 & 98.59 & 14min & 496.61m  &  flying \T \\
Kendal \etal \cite{kendall2015posenet}  & 49 & 91.79 & 8min & 596.7m &  flying\\
\textbf{Wang \etal  \cite{wang2017deepvo}} & \textbf{0} & \textbf{100.0} & \textbf{5min}  &  \textbf{443.11m} & \textbf{flying}\\
MTRL & 31 & 95.09 & 10min & 631.55m &  flying \B \\
\hline %mountain results
\multicolumn{6}{|c|}{Snowy Mountain} \T\B \\
\hline

Bojarski \etal \cite{bojarski2016end}  & 0 & - & 6min  & 65.59m &  hovering \T \\
Kendal \etal \cite{kendall2015posenet} & 0 & -  & 5min& 125.62m &  hovering\\
Wang \etal  \cite{wang2017deepvo}  & 0 & - & 13min  & 6.02m & hovering\\
\textbf{MTRL}  & \textbf{27} & \textbf{97.25} & \textbf{20min} & \textbf{982.64m} &  \textbf{flying} \B \\
\hline %plain field results
\multicolumn{6}{|c|}{Plain Field} \T\B \\
\hline

Bojarski \etal \cite{bojarski2016end} & 0 & - & 20min & 0.0m & hovering \T \\
Kendal \etal \cite{kendall2015posenet} & 0 & -  & 20min  & 0.0m &  hovering\\
Wang \etal  \cite{wang2017deepvo} & 0 & - & 15min & 0.0m &  hovering\\
\textbf{MTRL}  &  \textbf{7} & \textbf{99.19} & \textbf{11min} & \textbf{867.45m}  & \textbf{flying} \B \\

\hline
\end{tabular}
\label{table:table-1}
 }
\end{center}
%\todo[inline]{resize the table to fit to linewidth}

\end{table}

%\vspace{-0.5cm}
 \begin{figure}
    \centering
    \includegraphics[width=\linewidth]{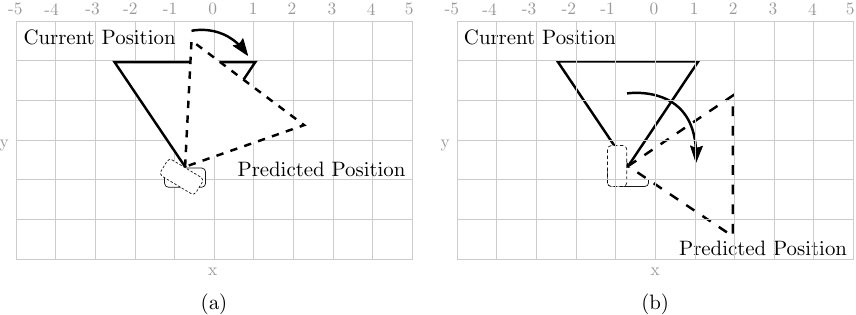}
    \vspace{-0.5cm}
    \caption{Illustration of the angular rotation of the UAV when the values in the $y$ position are close to zero (a), and far from zero (b).}
    \label{fig:fov}\vspace{-0.5cm}
    %\todo[inline]{Figure and Fonts are too small. Convert the figure to PDF}
\end{figure}

 \begin{figure*}
    \centering
    \includegraphics[width=\linewidth]{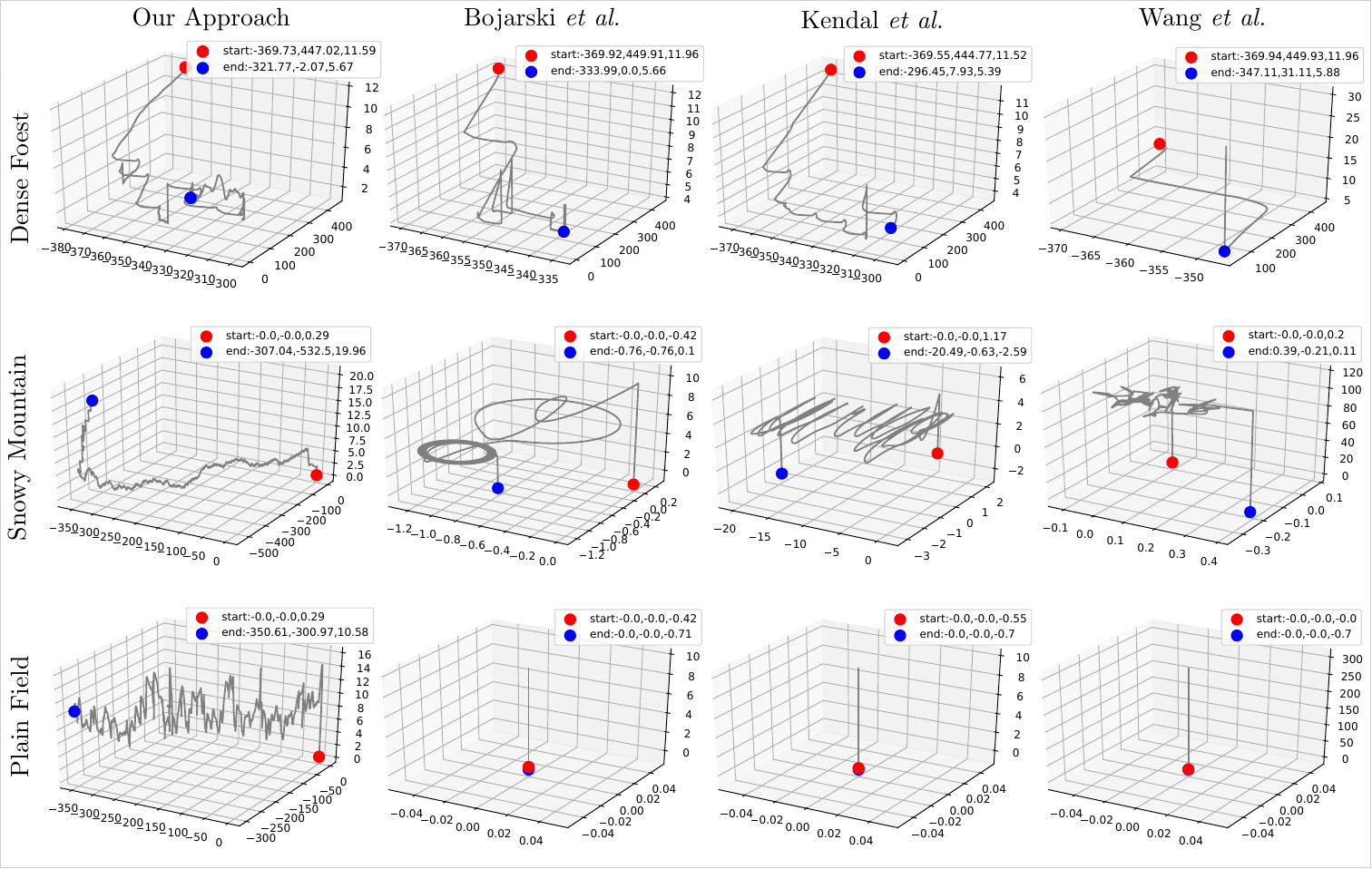}
    \vspace{-0.5cm}
    \caption{Comparison of each approach when autonomously flying under the canopy of a dense forest, over a snowy mountain and over a plain field.}
    \label{fig:test_stl}\vspace{-0.5cm}
    %\todo[inline]{Again, remove the whitespace around the figure.}
\end{figure*}

 From Figure \ref{fig:fov}, it can be observed that the higher the variance in the $y$ coordinate, the wider the FoV (Field of View) will be. At a later stage, the imagery gathered during navigation can be used to identify any object/person/animal of interest. As such, a wider FoV will undoubtedly be more favourable. Based on the qualitative results in Figure \ref{fig:gt_dist}, we can observe that the Bojarski \etal \cite{bojarski2016end} and Kendal \etal \cite{kendall2015posenet} approaches have a lower variance closer to zero for $y$; %\textbf{and in the case of Kendal \etal \cite{kendall2015posenet} approach the network is prone to rotate to the left direction, more often than to the right}. 
 contrarily, the proposed approach tends to forecast positional values in $y$ far from zero, which results in a wider angular rotation of the head.

In summary, it can be observed that the proposed approach performs better due to its ability to carry a high-density exploration of the search perimeter by predicting waypoints that leads to the navigation of different routes that are closer to each other. Besides, due to its angular rotation, the FoV of the proposed approach is significantly wider as compared to comparators.

\subsection{Behaviour}

During autonomous flight tests using the SITL, we observe that although the approach by Wang \etal  \cite{wang2017deepvo} has the highest reliability rate, which makes it quantitatively better than the other approaches (Table \ref{table:table-1}), the fact remains that it has, the worst performance when qualitatively analysing the flight mission in the dense forest environment (Figure \ref{fig:test_stl}). The network fails to learn exploratory behaviour, which causes the model to predict very similar values, thus resulting in the UAV constantly moving forward in a path. Additionally, this approach \cite{wang2017deepvo} produces significantly lower changes in the $y$ direction, appointing to small FoV. Similar behaviour is also observed in the findings of the approaches adopted by Bojarski \etal \cite{bojarski2016end} and Kendal \etal \cite{kendall2015posenet}.

A secondary behaviour observed in \cite{bojarski2016end} is the constant attempts to gain altitude above the canopy. In order to correct this behaviour, a function is created that regulates the altitude during the flight, which forces the UAV to remain under the canopy. Within real-world scenarios, UAV flights are commonly monitored by a Geo-Fencing mechanism \cite{pratyusha2013geo}. In contrast, the proposed approach traverses the greatest distance (631.55m) with significant changes in $y$, which indicates a greater FoV. Additionally, Figure \ref{fig:test_stl}, illustrates that the proposed approach has a precise exploratory behaviour under the canopy, characteristic of low-altitude flight and constant changes in the $y$ direction.  

Conventionally, sensor filtering, mechanical dampers and dynamic $g$ compensation are often used to reduce the effect of motor/propeller vibration before translating attitude commands to motor commands. Although a smooth navigation is desirable for the purpose of this work, we remove the sensor filtering from the RCL which results in a jittering motion. This allows us to observe anomalies, such as overshooting or drifting. Since drifting is caused by a minor increase in rotation rate in a pair of motors, when a low pass filter is applied to it, the rotation rate changes and the resulting total force equals to the gravitational force. This implies that the UAV changes its behaviour from drifting to hovering.

A clear distinction between behaviours is imperative since there is a strong relationship between generalisation and an increase in rotational rates, as depicted in Figure \ref{fig:test_stl}. Greater generalisation capabilities lead to a heightened rotational rate and mobility, while lower generalisation capabilities result in rotational rate values closer to zero and no generalisation results in hovering.

\subsection{Generalisation} 
 
In assessing the ability of the approaches to generalise to unseen domains, we find that the approaches in \cite{bojarski2016end, kendall2015posenet, wang2017deepvo} fail to generalise in the snowy mountain as well as the plain field (Figure \ref{fig:test_stl}). Since the values predicted by these approaches are mostly close to zero, the behaviour observed during the UAV flight is hovering or slowly drifting due to the simulated air flow velocity rather than flying. The hovering behaviour is evidence in the results derived from the plain field, where values predicted using the approaches by Bojarski \etal \cite{bojarski2016end}, Kendal \etal \cite{kendall2015posenet} and Wang \etal \cite{wang2017deepvo} cause the UAV to remain at position ($x:0,y:0$); only change is in the altitude ($z$). In contrast, the proposed approach is capable of generalising in all tested environments and has the greatest flight distance (Table \ref{table:table-1}).

%\vspace{-1cm}
\begin{figure}
    \centering
    \includegraphics[width=\columnwidth]{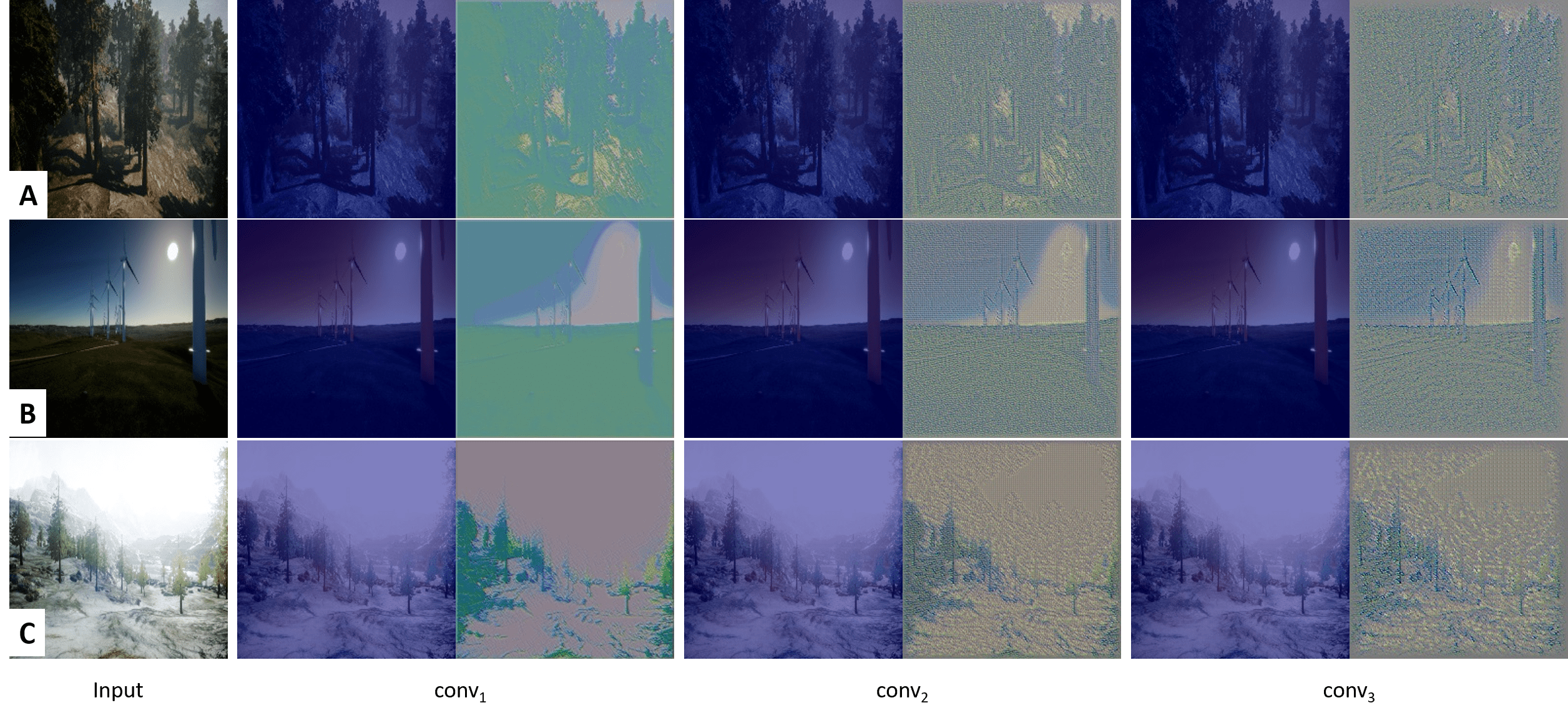}
    \vspace{-0.9cm}
    \caption{Heat and activation maps from the three last convolutional layers of the proposed approach, prior to flattening. Row (A) shows the results of testing the forest environment, while row (B) the plain field and (C) the snowy environment.}
    
    \label{fig:layers_mtl}
    %\todo[inline]{Fonts are too small. Convert the figure to PDF}
%\end{figure}
\vspace{0.2cm}
%\begin{figure}[H]
%    \centering
    \includegraphics[width=\columnwidth]{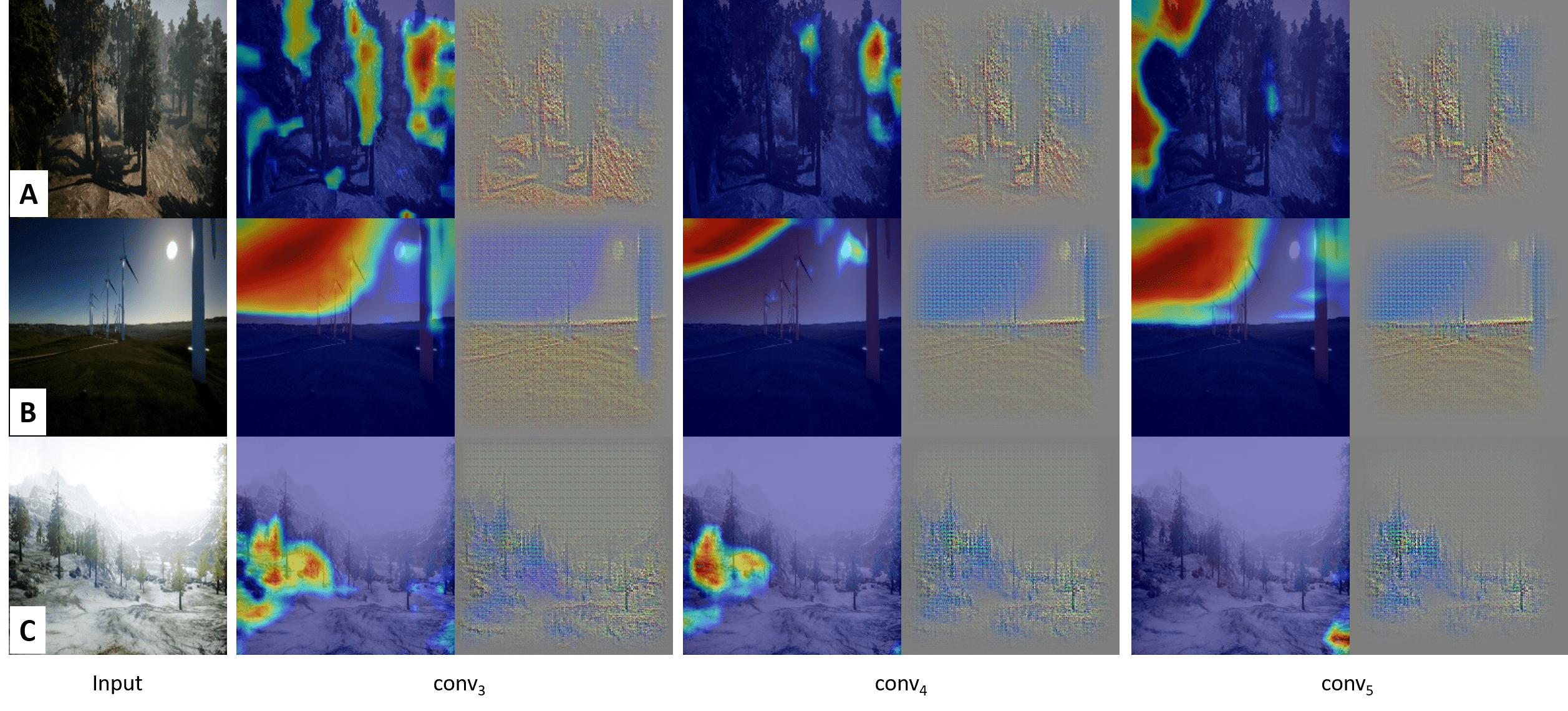}
    \vspace{-0.8cm}
    \caption{Three last convolutional layers of Bojarski \etal \cite{bojarski2016end} approach.}
    \label{fig:layers_nvidia}
%\end{figure}
\vspace{0.2cm}
%\begin{figure}[H]
%    \centering
    \includegraphics[width=\columnwidth]{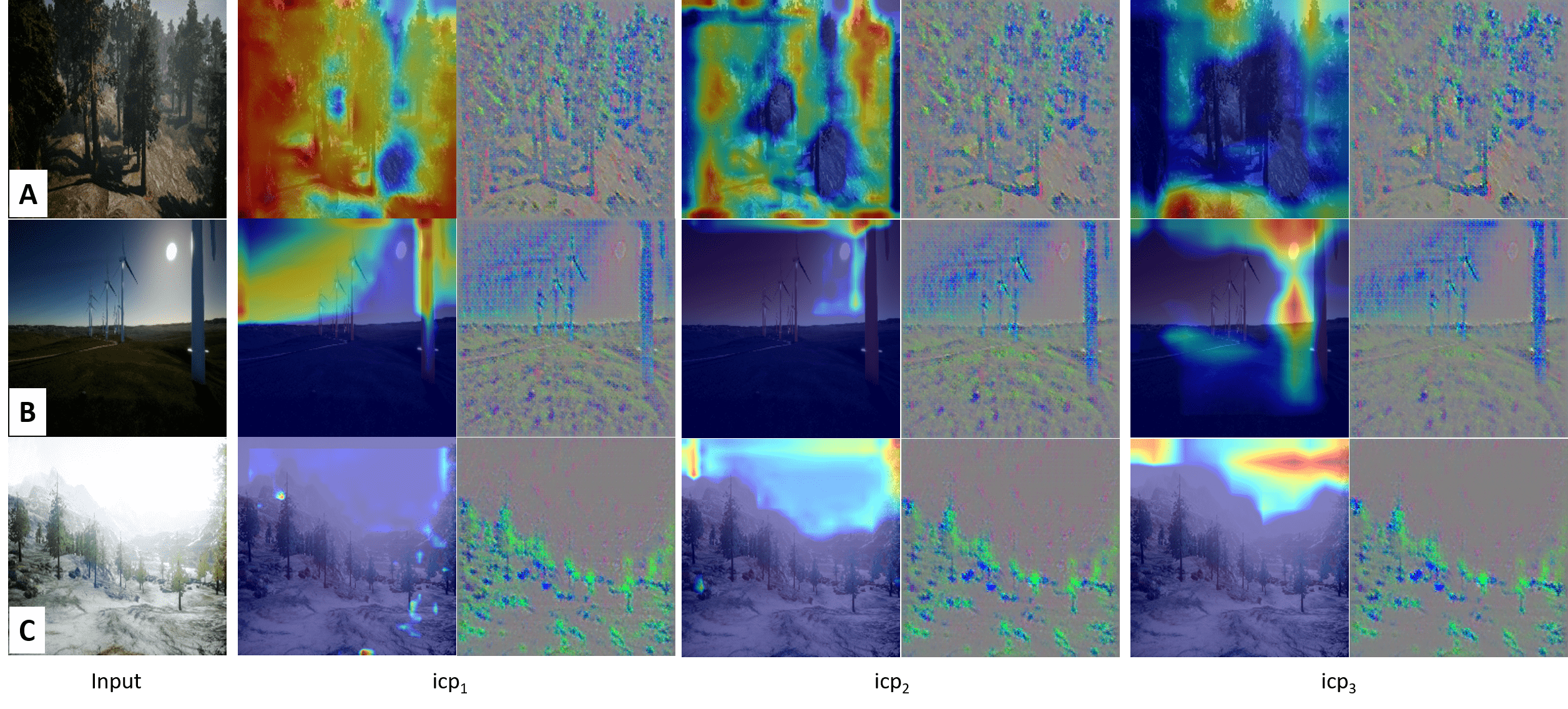}
    \vspace{-0.8cm}
    \caption{Three last inception layers of Kendall \etal \cite{kendall2015posenet} approach.}
    \label{fig:layers_posenet}
%\end{figure}
\vspace{0.2cm}

%\begin{figure}[H]
 %   \centering
    \includegraphics[width=\columnwidth]{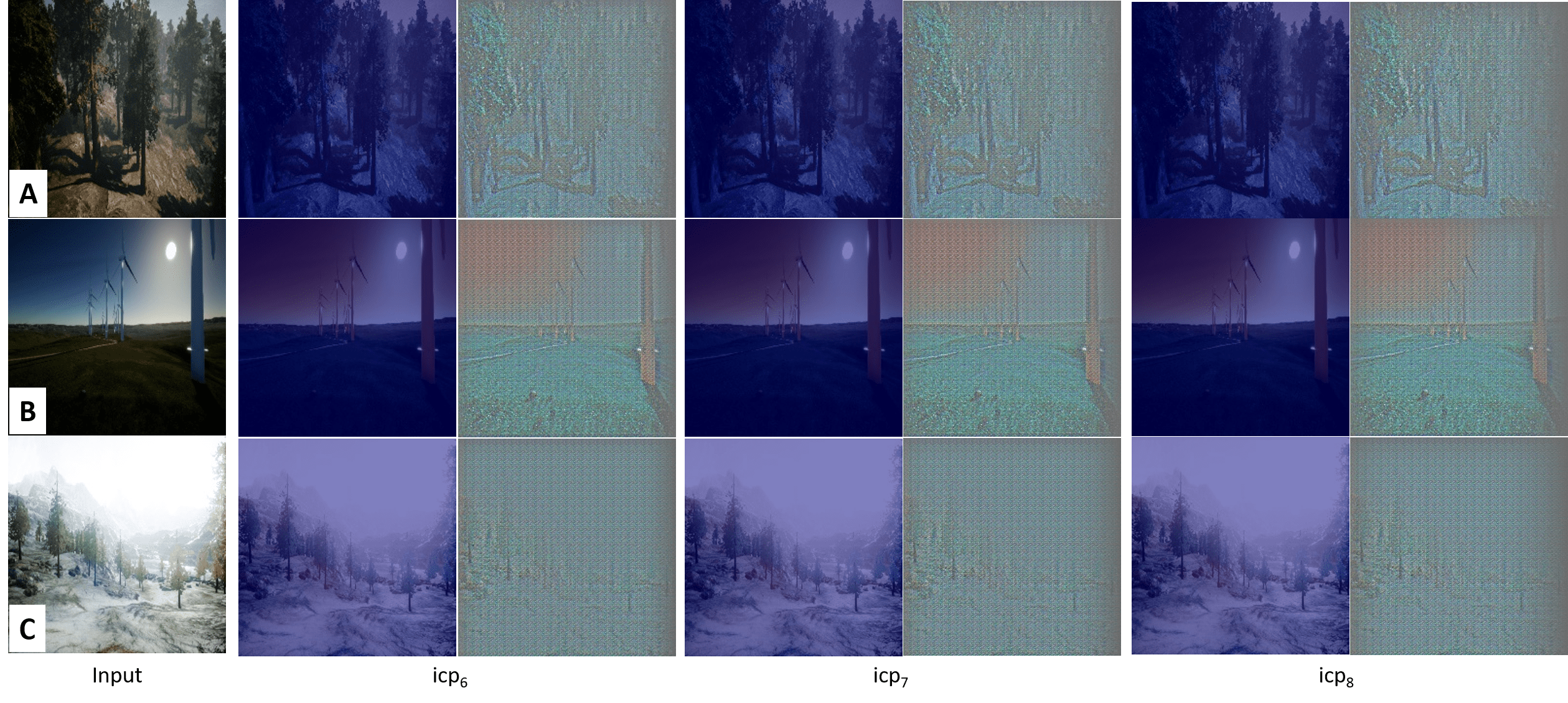}
    \vspace{-0.7cm}
    \caption{Three last convolutional layers of Wang \etal \cite{wang2017deepvo} approach.}
    \label{fig:layers_deepvo}
    \vspace{-0.5cm}
\end{figure}

The generalisation capability of the proposed approach can primarily be attributed to the shallow architecture of its network, in which learning is confined to local features that are commonly found in various obstacles from the global structure of the scene, such as edges and depth information (Figure \ref{fig:layers_mtl}). Consequently, the proposed approach does not learn to differentiate between a tree and a rock or a branch; instead, it learns to differentiate salient objects that are to be avoided from any other elements within the scene. When tested in a new environment, our model will try to avoid areas rich in salient obstacles/object features, regardless of what this saliency may entail.

As observed by the heat map and activation map in Figures \ref{fig:layers_nvidia}-\ref{fig:layers_deepvo}, the deeper the network, the more specific is the knowledge acquired about the training environment. This phenomenon is mainly attributed to the fact later layers tend not only to retain spatial information but also to learn high-level semantic information about the scene \cite{richter2017safe},\cite{selvaraju2017grad}, as is the case for the approaches by Kendall \etal    \cite{kendall2015posenet} (Figure \ref{fig:layers_posenet}) and Bojarski \etal \cite{bojarski2016end} (Figure \ref{fig:layers_nvidia}). In both these cases, the network highlights arbitrary regions within the heat map, which illustrate the origins of the features observed in the activation map. Here, blue areas signify lower certainty about the classification of the area of interest (ROI), while red areas denote higher certainty.  Furthermore, all three approaches \cite{kendall2015posenet, bojarski2016end, wang2017deepvo} suggest that, the last activation map for both the plain field and snowy mountain are smoother than the activation map for the forest environment, implying that the network is less certain regarding to its predictions in these environments than the forest environment \cite{richter2017safe}. Contrarily, in our proposed approach, we can observe an equal representation of salient features across all three environments, which is indicative of the superior generalisation capabilities of our approach.

Put succinctly, learning very specific details about the content of a given environment can prove to be very useful when navigating within the same environment or those with close similarity in their appearance. However, this can preclude generalisation to unseen environments, which are far more likely to be encountered in real-world applications. Although all comparators \cite{kendall2015posenet, bojarski2016end, wang2017deepvo} suffer from this predicament, our technique demonstrates significantly superior generalisation capabilities.

\subsection{Distance and Reliability}
Further experiments are also carried out to verify the validity of our approach when flying from different starting points within the forest and for a longer duration of time (Table \ref{table:table-2}). 
 
 \begin{table}[ht]
% \small
\caption{The multi-task approach in 6 distinct flight missions.}

\label{table:2}
\vspace{-0.5cm}
\begin{center}
\resizebox{\linewidth}{!}{%
\begin{tabular}{|l|c|c|c|c|}
\hline
Method  & NI & Reliability & Duration & Distance (m) \T\B \\
\hline\hline
MTRL (F1) & 2 & 97.46 & 4min  & 78.88m \T \\
\textbf{MTRL (F2)} & \textbf{21} & \textbf{98.07} & \textbf{9min}  & \textbf{1086.62m}\\
MTRL (F3) & 13 & 98.01 & 6min & 654.76m \\
MTRL (F4)  & 31 & 95.09 & 10min & 631.55m \\
\hline
MTRL (F5) & 197 & 90.55 & 2h40min & 2086.61m \\
MTRL (F6)  & 399 & 88.55 & 5h37min & 2287.37m \B \\

\hline
\end{tabular}
\label{table:table-2}
}
%\vspace{-0.5cm}
\end{center}
\end{table}

The results of the experiments presented in Table \ref{table:table-2} provide a better numerical evaluation of the robustness of the proposed approach. The set of tests F1-F4 have a navigational loop of 150 iterations, while tests F5 and F6 have a navigational loop of 1000 and 2000 iterations respectively. Experiments F4-F6 are premised on the initial starting position with coordinates of (70,-450, -12), while F1-F3 use the default starting position of (0,0,0) within the map. The difference concerning the time taken to complete each one of the flight mission F1-F3 is caused by choice of the route and the amount of airflow to overcome. Regardless of the distance or initial starting position, the level of reliability for each test is above 88\%.

    \section{Conclusion}
In this work, we present a deep learning approach based on Multi-task Regression-based Learning, which takes advantage of the use of shallow networks by learning each task individually. Experiments indicate that our method is capable of generalising to unseen domains and has a larger coverage area than comparators \cite{bojarski2016end, kendall2015posenet, wang2017deepvo}. Our method demonstrates a more aggressive exploratory behaviour due to a wider FoV in comparison with other approaches. Additionally, our techniques is particularly suitable for search and rescue operations in any above-ground environment, as it is does not require distinct pathways or GPS for navigation. An interesting direction for future research would be the investigation of the efficacy of the proposed approach when it is deployed indoors and in areas with limited navigational space, as well as boosting the performance of the approach by incorporating the task of depth estimation into overall multi-task model.

\bibliographystyle{IEEEtran}
\bibliography{ref/bibliography.bib}
\end{document}